# Can Neural Networks Count Digit Frequency?

**Padmaksh Khandelwal**

Sir Padampat Singhania School, Kota

ABSTRACT

In this research, we aim to compare the performance of different classical machine learning models and neural networks in identifying the frequency of occurrence of each digit in a given number. It has various applications in machine learning and computer vision, e.g. for obtaining the frequency of a target object in a visual scene. We considered this problem as a hybrid of classification and regression tasks. We carefully create our own datasets to observe systematic differences between different methods. We evaluate each of the methods using different metrics across multiple datasets.The metrics of performance used were the root mean squared error and mean absolute error for regression evaluation, and accuracy for classification performance evaluation. We observe that decision trees and random forests overfit to the dataset, due to their inherent bias, and are not able to generalize well. We also observe that the neural networks significantly outperform the classical machine learning models in terms of both the regression and classification metrics for both the 6-digit and 10-digit number datasets. Dataset and code are available on [github](github).

## Introduction

Some of the fundamental aspects of deep learning were introduced quite early, e.g. backpropagation[1] and deep convolutional neural networks[2], however, it required an increase in computational power and access to large datasets[3-5] to get mainstream. Recently, these learning techniques have been shown to be successful in different tasks like playing the game of Go[6] and even the task of question-answering interactions, e.g. instructGPT[7] which led to recently popular ChatGPT.

In this paper, we show that it is still not easy to use the recent machine learning models for a simple but important task of counting the frequency of different digits in a given sequence of numbers, e.g. Figure 1 shows that even ChatGPT is not good at this task. This task has several downstream applications, e.g. counting the number of objects detected in a scene[8,9]. We compare different classical machine learning and neural network-based methods for this task. As part of classical methods, we utilize decision trees[10,11] and random forests[12-14]. Thus, in this research work, we try to understand classical machine learning and neural network architectures and their effects.

**Figure 1**: Output of ChatGPT for our problem statement. The frequency of each digit in 1101111 should be 0 occurring once, 1 occurring 6 times, and the rest of the digits occurring 0 times.

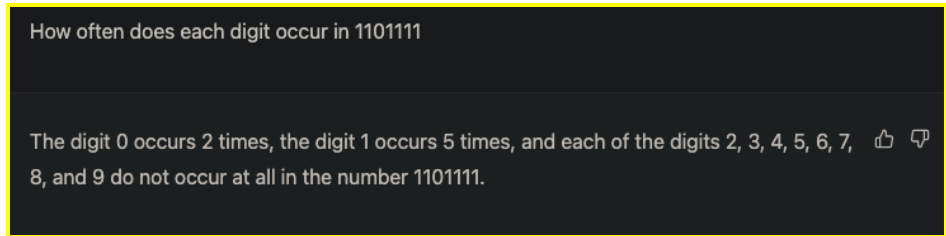

Decision Tree and Random Forests: A decision tree is created using a binary split which decides the branch to allocate for a data sample. The quality of a split is decided by a measure of impurity, e.g. "gini", which can be similar to the sum of the standard deviation of samples lying on each side of the split[15,16], hence the best split is likely to have the least "gini" score. Refer to Figures 6 to 9 to see decision tree structures. Decision trees can face the issue of overfitting which can be avoided by using random forests[12-14]. The basic idea behind random forests is to create lots of large decision trees such that their predictions are uncorrelated[14] and then take the average of their predictions, which is also called bagging[9]. There are different approaches to create uncorrelated models, e.g. by training them on different subsets of data, by considering a random subset of columns for each split, etc[12-14]. Random forests have been shown to work quite well in practice, which is also evident from this work.

Our major contributions in this work are listed below:
- We systematically create our own datasets to bring out the differences in performance of different methods.
- We carefully split the datasets into training, validation and test sets to test the generalization capabilities of different methods across dataset sizes.
- For fair evaluation of the methods, we do multiple runs of each method to obtain statistical results. We also consider different metrics for both regression-based evaluation and accuracy-based evaluation.
- We also list specific examples to observe the overfitting behavior of decision trees and random forests which is not observed in the neural networks.
- We also perform hyper-parameter tuning of the neural networks and provide our observation as part of the ablation studies.

## Methodology

In this work we tackle the task of counting the frequency of different digits in a given number, as shown in Figure 2 and Figure 3.

Dataset

For this we created our own datasets. We created two different datasets using 6-digit and 10-digit numbers. Both datasets have 150,000 randomly generated numbers. To create the ground-truth count of each digit in the number, we used a Python code. We further split both datasets in a 60:20:20 ratio for training, validation, and test set. So the training set has 90,000 samples, the validation set has 30,000 samples, and the test set has 30,000 samples. The split was done to allow for the fine-tuning of the hyperparameters of the neural networks on the validation set which can later be tested on the unseen and unbiased test set, whose samples follow the same distribution as the training and validation set.

The training set of size 90,000 represents 9% of the total possible 6-digit numbers. This can help us understand the generalization of the performance of machine learning models to unseen 6-digit numbers. To further challenge the generalizability of the models and test their capabilities to learn from limited data, we also considered a 10-digit numbers dataset as a 90,000-sized training set represents only 0.0009% of the total possible 10-digit numbers. We show that this change in the fraction of seen dataset (from 9% to 0.0009%) has the least effect on the performance of the neural networks[1,2] as compared to the classical machine learning models[10-14].

Implementation

For the implementation of the different machine learning models, we extensively used Jupyter Notebooks with the *scikit learn*[17] and *fastai*[18] libraries. While *scikit learn*[17] has several built-in classical ML models, *fastai*[18] has implementations of several state-of-the-art deep learning models. Using these libraries help us overcome the challenge of tediously and manually assigning all hyperparameters and thus allows us to quickly experiment with multiple methods and techniques.

We decided to use the decision tree and random forest regressor as classical ML models. Decision trees[10] build regression or classification models in the form of a tree structure. At every node, it splits a dataset into two subsets such that the "gini" score is minimized, to incrementally develop the decision tree. The final result is a tree with decision nodes and leaf nodes. A random forest[13,14] is a meta-estimator that fits a number of classifying decision trees on various sub-samples of the dataset and uses averaging to improve the predictive accuracy and avoid over-fitting.

**Figure 2**: 6-Digit Original Dataset: a sequence of 6-digit number (rightmost column) and the corresponding count of each digit

| | Count of digit 0 | Count of digit 1 | Count of digit 2 | Count of digit 3 | Count of digit 4 | Count of digit 5 | Count of digit 6 | Count of digit 7 | Count of digit 8 | Count of digit 9 | Number |
|---|---|---|---|---|---|---|---|---|---|---|---|
| 0 | 0 | 1 | 2 | 1 | 0 | 1 | 0 | 1 | 0 | 0 | 175322 |
| 1 | 0 | 0 | 1 | 0 | 0 | 0 | 1 | 0 | 1 | 3 | 989296 |
| 2 | 0 | 0 | 1 | 0 | 2 | 0 | 0 | 2 | 0 | 1 | 977424 |
| 3 | 3 | 1 | 0 | 1 | 0 | 0 | 0 | 0 | 1 | 0 | 100083 |
| 4 | 1 | 1 | 1 | 1 | 0 | 0 | 0 | 1 | 0 | 1 | 921037 |
| ... | ... | ... | ... | ... | ... | ... | ... | ... | ... | ... | ... |
| 149995 | 1 | 1 | 0 | 1 | 1 | 1 | 0 | 0 | 0 | 1 | 105439 |
| 149996 | 1 | 0 | 0 | 1 | 0 | 0 | 1 | 1 | 0 | 2 | 679039 |
| 149997 | 0 | 1 | 0 | 1 | 0 | 1 | 2 | 0 | 1 | 0 | 581366 |
| 149998 | 0 | 1 | 0 | 0 | 0 | 1 | 2 | 1 | 0 | 1 | 796615 |
| 149999 | 1 | 0 | 0 | 0 | 1 | 0 | 1 | 1 | 1 | 1 | 790864 |

150000 rows × 11 columns

**Figure 3**: 10-Digit Original Dataset: a sequence of 10-digit number (rightmost column) and the corresponding count of each digit

| | Count of digit 0 | Count of digit 1 | Count of digit 2 | Count of digit 3 | Count of digit 4 | Count of digit 5 | Count of digit 6 | Count of digit 7 | Count of digit 8 | Count of digit 9 | Number |
|---|---|---|---|---|---|---|---|---|---|---|---|
| 0 | 0 | 3 | 0 | 1 | 3 | 0 | 2 | 0 | 1 | 0 | 6131146484 |
| 1 | 2 | 2 | 0 | 2 | 0 | 0 | 0 | 2 | 1 | 1 | 7911830037 |
| 2 | 0 | 1 | 1 | 2 | 2 | 2 | 0 | 0 | 0 | 2 | 3453425199 |
| 3 | 0 | 0 | 0 | 2 | 3 | 1 | 0 | 2 | 2 | 0 | 5438747348 |
| 4 | 1 | 3 | 1 | 2 | 1 | 0 | 1 | 0 | 0 | 1 | 4012163319 |
| ... | ... | ... | ... | ... | ... | ... | ... | ... | ... | ... | ... |
| 149995 | 3 | 1 | 0 | 0 | 1 | 1 | 1 | 0 | 2 | 1 | 5409016808 |
| 149996 | 1 | 1 | 2 | 1 | 2 | 1 | 0 | 0 | 1 | 1 | 1430982524 |
| 149997 | 1 | 1 | 0 | 1 | 0 | 0 | 1 | 2 | 2 | 2 | 9901376878 |
| 149998 | 1 | 0 | 1 | 1 | 2 | 1 | 0 | 2 | 2 | 0 | 2875473048 |
| 149999 | 1 | 0 | 3 | 0 | 0 | 0 | 0 | 2 | 4 | 0 | 2288082877 |

150000 rows × 11 columns

The dataset follows a specific labeling pattern, hence we believe that the decision tree could, perhaps, identify the necessary comparisons to perfectly, or nearly perfectly, predict the pattern. Random forest in general is the best performing and the most versatile classical ML model and is a key reason for its widespread popularity and, thus, also stood out as a possibly strong baseline.

Let $x_i$ be the $i^{th}$ number or sample for $1 \leq i \leq n$, let $y_i$ be the ground-truth label vector for the $i^{th}$ number such that $y_{ij}$ is the count of $j^{th}$ digit for $0 \leq j \leq 9$, and $\hat{y}_i$ be the predicted vector for the $i^{th}$ number such that $\hat{y}_{ij}$ is the count of $j^{th}$ digit for $0 \leq j \leq 9$.

The regression performance metrics we consider are root mean squared error and mean absolute error, the two popular metrics in regression, and the classification metric we consider is accuracy. Root mean squared error is calculated as

$$RMSE = \sqrt{\sum_{i=1}^{n} \sum_{j=0}^{l-1} \frac{(y_{ij} - \hat{y}_{ij})^2}{nl}}$$

and the mean absolute error is calculated as

$$MAE = \sum_{i=1}^{n} \sum_{j=0}^{l-1} \frac{|y_{ij} - \hat{y}_{ij}|}{nl}$$

, where $n$ is the total number of samples (or numbers), $l$ is the length of output vector (which is 10 for the count of 10 digits), $y_i$ is the $i^{th}$ ground-truth label vector; and $\hat{y}_i$ is the $i^{th}$ predicted vector.

The problem statement can be tackled either using a regression method or classification method. The count of each of the 10 digits is only limited to integers 0 to 6 for the 6-digit set and 0 to 10 for the 10-digit set. However, if we consider a classification method, the presence of different digits would require an excessively complex and yet underperforming multi-class multi-label classification method which may easily overfit the small fraction of real data we have.

Therefore, to tackle this problem, we first implemented multi-class regression models and generated the two error metrics and, then modified the predictions to be rounded off to the nearest whole number (predictions less than zero rounded up to zero and those more than the total number of digits rounded down to the total digits themselves (6 and 10 respectively). We can therefore also consider accuracy metric over these predictions which we define as:

$$Accuracy = \sum_{i=1}^{n} \sum_{j=0}^{l-1} \frac{I(y_{ij} = \hat{y}_{ij})}{nl}$$

where $I(.)$ is an indicator function which is 1 when $y_{ij} = \hat{y}_{ij}$, otherwise 0. For this, a program was made to check if each count of digits matched with the actual count and assigned it a 'one' if the count matched correctly, and 'zero' otherwise.

For example, if the $i^{th}$ number is $x_i$=153,236, the corresponding ground-truth label vector is $y_i$ =[0,1,1,2,0,1,1,0,0,0], representing the counts of digits 0 to 9, and the predicted vector is $\hat{y}_i$ =[0,0,1,2,0,0,1,0,0,0], then the accuracy of the method on the $i^{th}$ sample is

$$Accuracy = \sum_{j=0}^{l-1} \frac{I(y_{ij} = \hat{y}_{ij})}{nl} = \frac{8}{10} = 0.8$$

where $I(.)$ is an indicator function which is 1 when $y_{ij} = \hat{y}_{ij}$, otherwise 0.

*Modified Dataset*

Finally, it was discovered that the single feature dataset (the number itself as the only independent column) could be improved on by increasing the number of feature columns; we implemented it by placing each digit of the original number into a separate input column and removing the "Number" column, resulting into 6 and 10 independent columns respectively for the 6-digit and 10-digit datasets, refer Figure 4 and Figure 5. This had a small but noticeable improvement in the decision tree's performance[10,11] and a substantially larger improvement in random forest's performance[12,13]. For the neural networks, we only used this modified dataset.

**Figure 4**: 6-Digit Original Dataset with 16 columns: a sequence of 6-digit (rightmost 6 columns) and the corresponding count of each digit (left columns)

| | Count of digit 0 | Count of digit 1 | Count of digit 2 | Count of digit 3 | Count of digit 4 | Count of digit 5 | Count of digit 6 | Count of digit 7 | Count of digit 8 | Count of digit 9 | Digit 1 | Digit 2 | Digit 3 | Digit 4 | Digit 5 | Digit 6 |
|---|---|---|---|---|---|---|---|---|---|---|---|---|---|---|---|---|
| 0 | 0 | 1 | 2 | 1 | 0 | 1 | 0 | 1 | 0 | 0 | 1 | 7 | 5 | 3 | 2 | 2 |
| 1 | 0 | 0 | 1 | 0 | 0 | 0 | 1 | 0 | 1 | 3 | 9 | 8 | 9 | 2 | 9 | 6 |
| 2 | 0 | 0 | 1 | 0 | 2 | 0 | 0 | 2 | 0 | 1 | 9 | 7 | 7 | 4 | 2 | 4 |
| 3 | 3 | 1 | 0 | 1 | 0 | 0 | 0 | 0 | 1 | 0 | 1 | 0 | 0 | 0 | 8 | 3 |
| 4 | 1 | 1 | 1 | 1 | 0 | 0 | 0 | 1 | 0 | 1 | 9 | 2 | 1 | 0 | 3 | 7 |
| ... | ... | ... | ... | ... | ... | ... | ... | ... | ... | ... | ... | ... | ... | ... | ... | ... |
| 149995 | 0 | 1 | 2 | 0 | 1 | 0 | 1 | 1 | 0 | 0 | 6 | 7 | 1 | 2 | 4 | 2 |
| 149996 | 2 | 1 | 0 | 0 | 0 | 1 | 0 | 2 | 0 | 0 | 7 | 0 | 0 | 1 | 7 | 5 |
| 149997 | 0 | 0 | 1 | 1 | 1 | 1 | 0 | 1 | 0 | 1 | 3 | 4 | 9 | 7 | 5 | 2 |
| 149998 | 1 | 1 | 1 | 0 | 0 | 0 | 0 | 1 | 1 | 1 | 2 | 0 | 9 | 7 | 8 | 1 |
| 149999 | 0 | 0 | 0 | 0 | 0 | 1 | 1 | 3 | 1 | 0 | 5 | 8 | 7 | 6 | 7 | 7 |

150000 rows × 16 columns

**Figure 5**: 10-Digit Original Dataset with 20 columns: a sequence of 10-digit (rightmost 10 columns) and the corresponding count of each digit (left columns)

| | Count of digit 0 | Count of digit 1 | Count of digit 2 | Count of digit 3 | Count of digit 4 | Count of digit 5 | Count of digit 6 | Count of digit 7 | Count of digit 8 | Count of digit 9 | Digit 1 | Digit 2 | Digit 3 | Digit 4 | Digit 5 | Digit 6 | Digit 7 | Digit 8 | Digit 9 | Digit 10 |
|---|---|---|---|---|---|---|---|---|---|---|---|---|---|---|---|---|---|---|---|---|
| 0 | 3 | 1 | 1 | 0 | 1 | 0 | 0 | 4 | 0 | 0 | 7 | 4 | 1 | 2 | 0 | 7 | 7 | 0 | 7 | 0 |
| 1 | 0 | 2 | 1 | 1 | 1 | 0 | 1 | 4 | 0 | 0 | 1 | 7 | 4 | 6 | 3 | 7 | 7 | 1 | 2 | 7 |
| 2 | 1 | 1 | 3 | 0 | 1 | 2 | 2 | 0 | 0 | 0 | 2 | 1 | 2 | 4 | 5 | 2 | 6 | 0 | 5 | 6 |
| 3 | 1 | 2 | 2 | 1 | 0 | 0 | 1 | 2 | 1 | 0 | 1 | 3 | 7 | 1 | 2 | 6 | 8 | 0 | 2 | 7 |
| 4 | 0 | 2 | 1 | 1 | 1 | 0 | 0 | 1 | 3 | 1 | 1 | 8 | 9 | 8 | 2 | 3 | 8 | 4 | 7 | 1 |
| ... | ... | ... | ... | ... | ... | ... | ... | ... | ... | ... | ... | ... | ... | ... | ... | ... | ... | ... | ... | ... |
| 149995 | 1 | 0 | 2 | 0 | 2 | 0 | 1 | 1 | 2 | 1 | 8 | 4 | 6 | 2 | 4 | 0 | 8 | 2 | 9 | 7 |
| 149996 | 0 | 0 | 1 | 1 | 3 | 2 | 0 | 0 | 1 | 2 | 9 | 3 | 2 | 9 | 4 | 8 | 5 | 4 | 5 | 4 |
| 149997 | 3 | 0 | 0 | 1 | 1 | 0 | 3 | 0 | 1 | 1 | 8 | 4 | 0 | 6 | 6 | 3 | 0 | 9 | 6 | 0 |
| 149998 | 1 | 0 | 1 | 1 | 1 | 0 | 0 | 2 | 3 | 1 | 9 | 8 | 8 | 0 | 2 | 8 | 3 | 4 | 7 | 7 |
| 149999 | 1 | 0 | 0 | 1 | 2 | 1 | 2 | 0 | 1 | 2 | 6 | 6 | 9 | 9 | 3 | 0 | 4 | 8 | 4 | 5 |

150000 rows × 20 columns

All the neural networks were composed of input layers, dense linear layers, and dense non-linear layers, which implement ReLUs (Rectified Linear Units)[3] as activation functions, SGD[1-3], and Adam optimizers[19]. For reference, a ReLU layer is used to implement a non-linearity in the neural network to better trace a non-linear pattern, which is essentially an identity function for all non-negative values, and zero for negative values.

## Experiments

The results show that neural networks performed significantly better than the decision tree and random forest
models, especially when using the modified dataset. The best results were obtained by using the appropriate number of layers, learning rate, and number of epochs.

The results are shown in Tables 1, 2, 3, and 4. For reference, the following keys are provided to identify the different models:
- Decision Tree 1 - Decision Tree trained on the original dataset
- Random Forest 1 - Random Forest trained on the original dataset
- Decision Tree 2 - Decision Tree trained on the modified dataset
- Random Forest 2 - Random Forest trained on the modified dataset
- Neural Network - *fastai.tabular* implemented neural network[20]
- Neural Network + Embedding - *fastai.tabular* neural network implemented with a hidden embedding[20].

We report RMSE, MAE and Accuracy metrics for each of the methods. We run each method multiple times on the validation set to obtain statistical errors. The results are consistent for both the 6-digit and 10-digit datasets, and by employing both the regression and classification metrics. However, it is key to note that even the neural networks do not have perfect accuracy but it is almost 100%.

**Table 1**. 6-Digit Validation Set. For statistical error, each method was run 5 times.

| Method | RMSE | MAE | Accuracy |
| --- | --- | --- | --- |
| Decision Tree 1 | 0.523±0.000 | 0.253±0.000 | 90.206% |
| Decision Tree 2 | 0.516±0.000 | 0.249±0.001 | 75.888% |
| Random Forest 1 | 0.463±0.000 | 0.277±0.000 | 89.777% |
| Random Forest 2 | 0.281±0.001 | 0.213±0.001 | 92.886% |
| Neural Network | **0.178±0.015** | 0.136±0.015 | **98.957%** |
| Neural Network + Embedding | 0.180±0.015 | **0.135±0.010** | 98.366% |

**Table 2**. 6-Digit Test Set

| Method | RMSE | MAE | Accuracy |
| --- | --- | --- | --- |
| Decision Tree 1 | 0.522 | 0.253 | 90.201% |
| Decision Tree 2 | 0.516 | 0.250 | 75.861% |
| Random Forest 1 | 0.463 | 0.277 | 93.058% |
| Random Forest 2 | 0.281 | 0.215 | 93.015% |

| Method | RMSE | MAE | Accuracy |
|---|---|---|---|
| Neural Network | 0.181 | 0.134 | 99.440% |
| Neural Network + Embedding | 0.183 | 0.139 | 99.156% |

**Table 3**. 10-Digit Validation Set. For statistical error, each method was run 5 times.

| Method | RMSE | MAE | Accuracy |
|---|---|---|---|
| Decision Tree 1 | 0.997±0.000 | 0.693±0.000 | 44.167% |
| Decision Tree 2 | 1.021±0.001 | 0.714±0.000 | 42.994% |
| Random Forest 1 | 0.862±0.000 | 0.666±0.000 | 44.583% |
| Random Forest 2 | 0.623±0.001 | 0.499±0.001 | 53.019% |
| Neural Network | 0.293±0.025 | 0.221±0.018 | 98.256% |
| Neural Network + Embedding | 0.210±0.014 | 0.162±0.010 | 96.965% |

**Table 4**. 10-Digit Test Set

| Method | RMSE | MAE | Accuracy |
|---|---|---|---|
| Decision Tree 1 | 0.998 | 0.693 | 43.986% |
| Decision Tree 2 | 1.018 | 0.712 | 43.198% |
| Random Forest 1 | 0.864 | 0.666 | 44.545% |
| Random Forest 2 | 0.620 | 0.495 | 52.827% |
| Neural Network | 0.303 | 0.216 | 97.833% |
| Neural Network + Embedding | 0.274 | 0.208 | 97.920% |

Observations on Classical ML Models

<u>Generalization:</u> The performance of classical ML models is greatly affected by the number of digits, the RMSE and MAE nearly doubled, whereas accuracy halved. On the contrary, the neural networks are only slightly affected or nearly unaffected by the increase in the digits, especially considering the large difference in the proportionality of more possible values in 6-digit and 10-digit numbers as mentioned earlier.

Modified dataset effect: It is observed that the modified dataset improves the performance of both decision trees and random forests, however, substantially more for random forests. This could be attributed to the tendency of random forests to generate many decision trees over multiple different features, instead of a single feature which generated the one and only possible tree given in the figures below. The averaging process of random forests[12,13] over several decision trees in the modified dataset and on multiple batches of random, unbiased data is responsible for generating different outputs every time they are run and causing substantially less error and more accuracy compared to the performance on the original dataset.
This could also be the explanation for the decision trees and random forests generating exactly the same performance consistently on the original datasets for both 6-digit and 10-digit numbers across multiple runs, thus, having no change in the statistical error, as only a single decision tree is possible and only a single set of decision trees and their respective batches are being computed in the random forest.

Decision tree overfits: As we used decision tree analysis methods, it was observed that the decision tree had created over 85,000 leaf nodes for the training dataset of 90,000 numbers for both datasets, which is a clear example of an overfitting and memorizing model.
The random forest model performed slightly better than the decision tree model; however, it is worth mentioning that as a random forest creates many decision trees on unbiased data and bags them together, it will always outperform decision trees. It is also worth noting that the decision tree created many numerical splits to make nodes and for inference, it simply outputs the average of the count of each digit across numbers reaching a leaf node during training, refer to Figure 6, Figure 7, Figure 8 and Figure 9, which shows that both the classical ML models clearly could not interpret any patterns.

**Figure 6**: First 6 nodes of the decision tree for the original 6-digit training dataset, (a): the top part, and (b): the lower part of the decision tree.

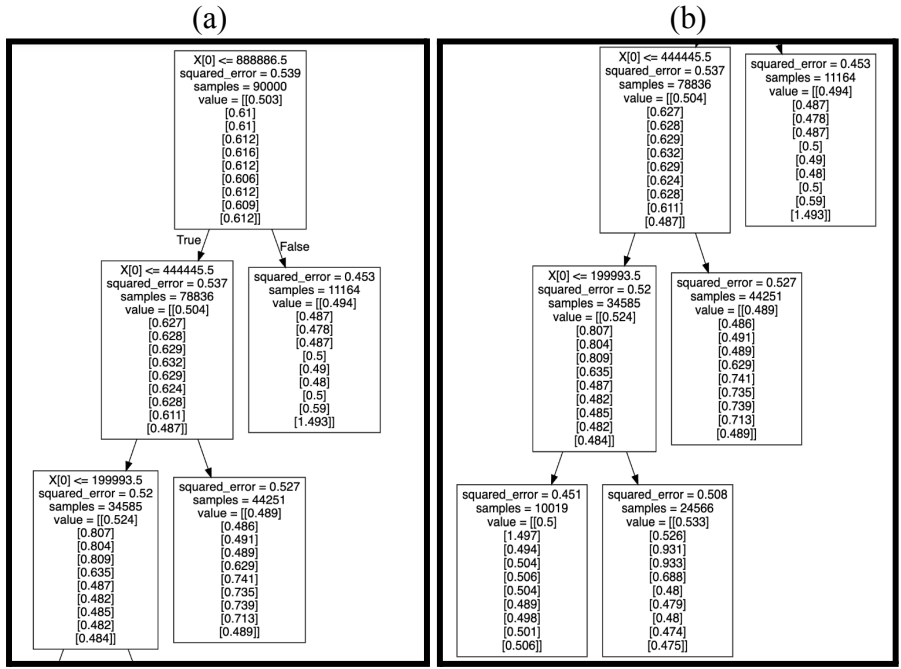

**Figure 7**: First 6 nodes of the decision tree for the modified 6-digit training dataset

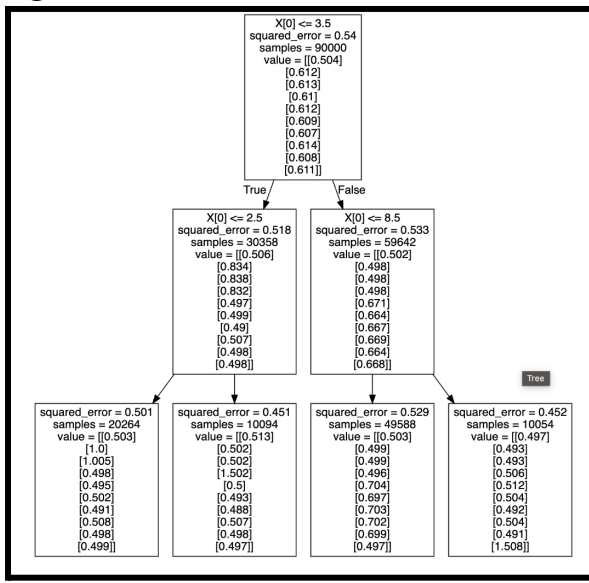

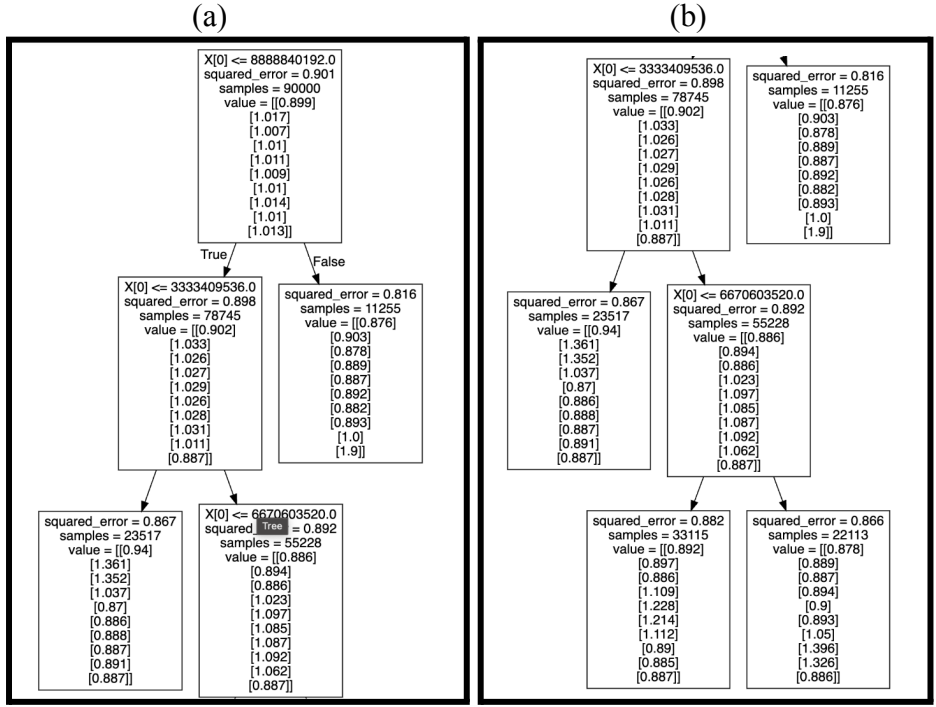

**Figure 8**: First 6 nodes of the decision tree for the original 10-digit training dataset (a): the top part, and (b): the lower part of the decision tree.

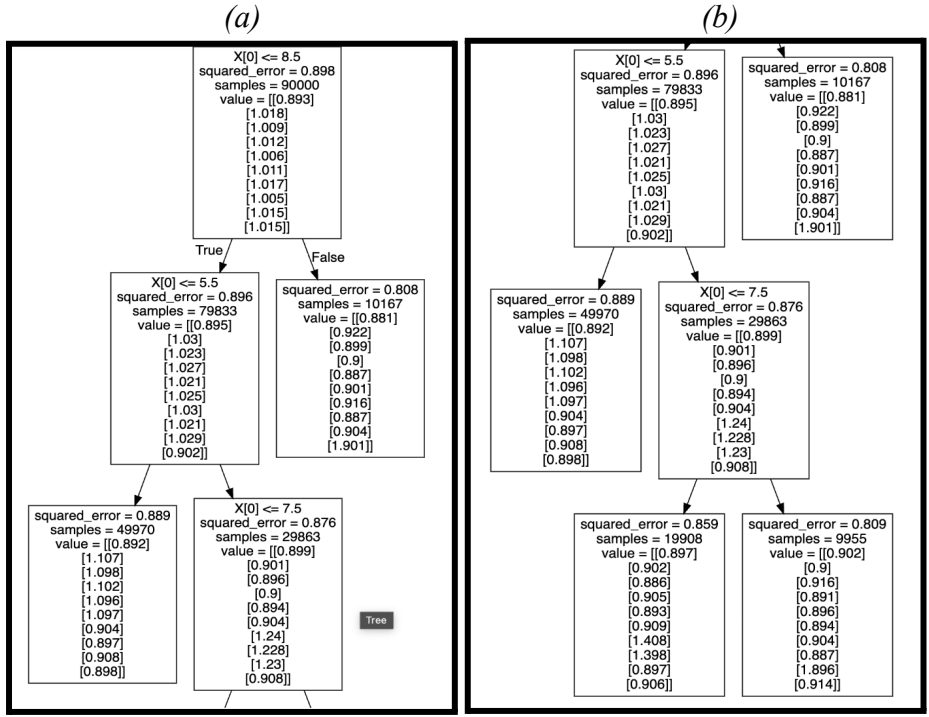

**Figure 9**: First 6 nodes of the decision tree for the modified 10-digit training dataset (a): the top part, and (b): the lower part of the decision tree.

*Some Special Cases*

We also experimented with a handful of outlier data points or numbers to observe predictions of the classical ML models.
For the original 6-digit dataset we tried the two pairs of consecutive numbers: (999998, 999999) and (100000, 100001). The decision tree predicted [0, 0, 0, 0, 1, 0, 0, 0, 0, 5] for both numbers of the first pair and [4, 2, 0, 0, 0, 0, 0, 0, 0, 0] for both numbers of the second pair, and random forest after undergoing the classification modification predicted [0, 0, 0, 0, 0, 0, 0, 1, 0, 5] for the first pair and [4, 2, 0, 0, 0, 0, 0, 0, 0, 0] for the second pair. Rerunning the classical ML models on the modified dataset still generated similar results: the decision tree predicted [0, 0, 0, 0, 1, 0, 0, 0, 0, 5] for the first pair and [3, 3, 0, 0, 0, 0, 0, 0, 0, 0] for the second pair, and random forest after undergoing the classification modification predicted [0, 0, 0, 0, 0, 0, 0, 1, 1, 5] for first pair and [3, 3, 0, 0, 0, 0, 0, 0, 0, 0] for the second. Thus these classical methods are making the same prediction for the successive numbers. This shows the inherent limitation of the decision tree and random forest, as they are splitting the nodes based on the numeric values of the numbers and not the count of each digit.

For the 10-digit dataset, we tried the two pairs of numbers: (9999999999, 9999999998) and (1000000000, 1000000001). The decision tree predicted [0, 1, 1, 0, 2, 0, 0, 0, 0, 6] for the former and [4, 2, 0, 1, 2, 0, 0, 0, 1, 0] for the latter. The random forest, whereas, predicted [0.02, 0.61, 0.71, 0.26, 1.31, 0.2, 0.29, 0.35, 0.75, 5.5] for the former and [3.57, 1.67, 0.52, 0.95, 1.81, 0.05, 0.4, 0.02, 0.57, 0.44] for the latter which after the classification modification are [0, 1, 1, 0, 1, 0, 0, 0, 1, 6] and [4, 2, 0, 1, 2, 0, 0, 0, 1, 0] respectively. The results are similar for the modified dataset. Evidently, this is another indication of the memorization that these classical ML models underwent and how they failed poorly in pattern recognition, which is even more evident in the 10-digit dataset.

Observations on Neural Networks

The neural networks, as aforementioned, outperformed classical ML models in every scenario and for both datasets. According to our hyperparameter optimization, we found the following best values for all the different scenarios using 16 epochs and [x,y,z] layers, where x,y, and z respectively are the number of parameters in each of the non-linear (ReLU [6]) hidden layers:
- 6-Digit numbers
  - Neural Network - Layers = [96,96,96], Learning Rate = 0.01
  - Neural Network with Embedding - Layers = [96,96,96], Learning Rate = 0.01, Embeddings are [10,100] by considering each of the 10 unique digits
- 10-Digit numbers
  - Neural Network - Layers = [128,128,128], Learning Rate = 0.01
  - Neural Network with Embedding - Layers = [256,256,256], Learning Rate = 0.005, Embeddings are [10,100] by considering each of the 10 unique digits

It could be hypothesized that as the neural networks utilize stochastic gradient descent to minimize loss by altering the parameters or weights and implement non-linearities through the ReLU layers, they at least trace out the non-linear pattern very well[1,2]. The 100-dimensional embeddings were used as an input feature for each of the ten possible values. Overall they did not significantly alter the predictions across the different metrics.

It is an intriguing detail that the classical ML models, which gave an accuracy of nearly 90% for 6-digit numbers, although by memorization, fell to less than or nearly 50% accuracy for 10-digit ones. On the contrary, neural networks hardly changed by even 1% in accuracy across datasets. They also produced less than half the errors compared to the best classical ML model baseline, which is the random forest, in both metrics. The following loss curve vs the number of epochs graphs, refer to Figure 10(a), 10(b), 10(c) and 10(d), indicate that the neural networks did not undergo any form of overfitting or memorization. This shows the generalization capability of neural networks.

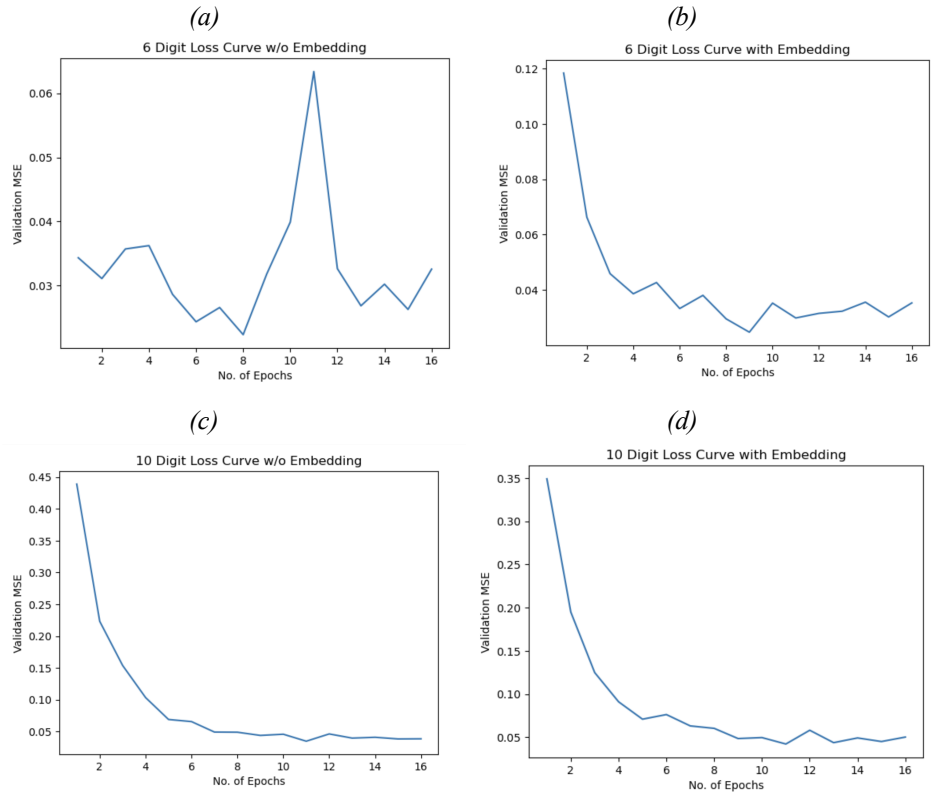

**Figure 10. (a)-(d):** *Loss (MSE) Curves for Neural Networks vs Number of Epochs*

*Some Special Cases And Comparisons With ChatGPT*

Similar to the classical ML models, we also worked with the following consecutive numbers for the neural networks: 6-digit numbers – (999999, 999998) and (100000, 100001); 10-digit numbers – (9999999999, 9999999998) and (1000000000,1000000001). Here are firstly the results by ChatGPT3 when asked for the task for recognizing the frequency of each digit in the above numbers, refer to Figure 11(a), 11(b), 11(c), 11(d), 11(e), 11(f).

**Figure 11. (a) - (f):** *ChatGPT3 responses for the above-mentioned numbers*

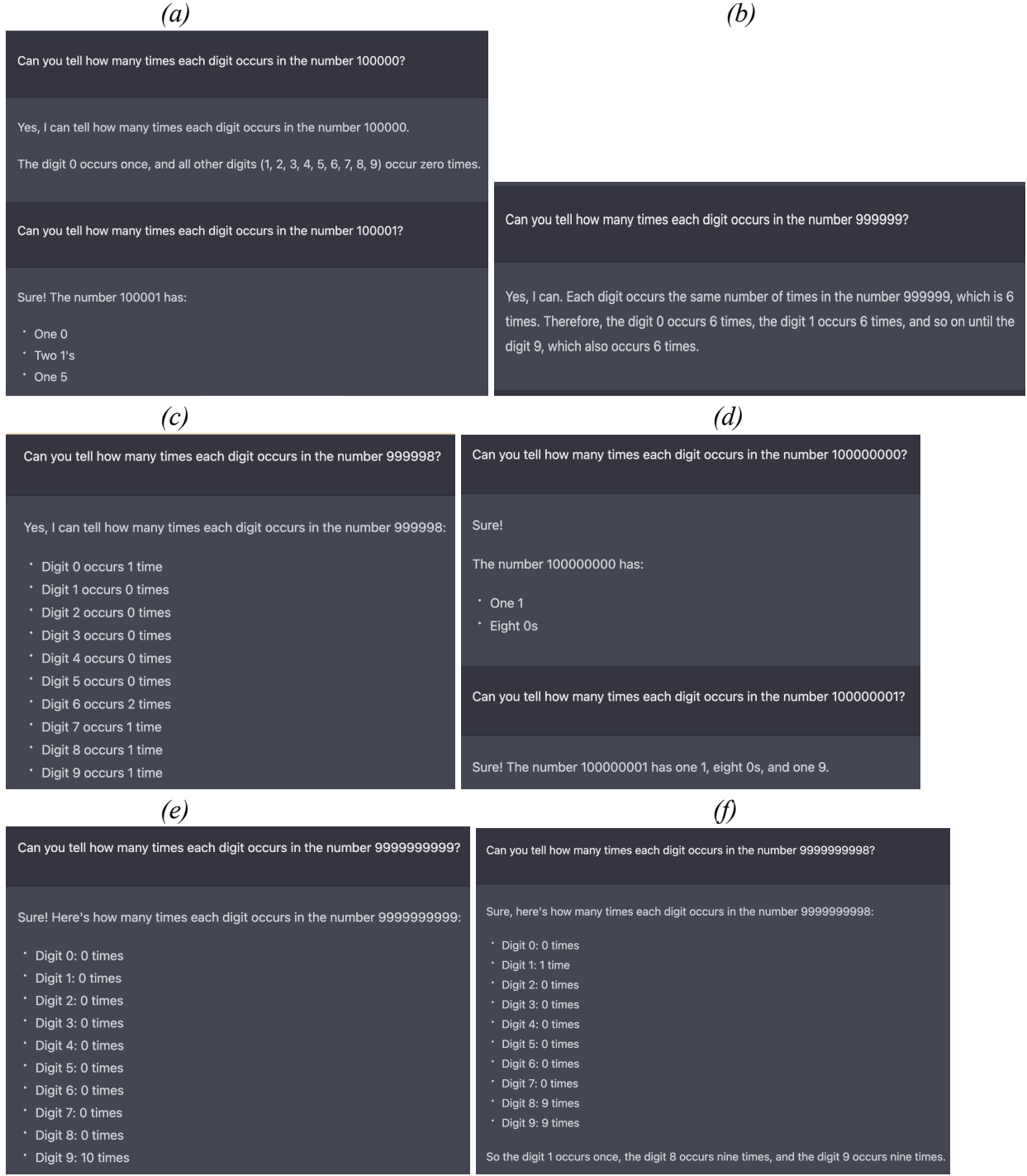

To summarize the results, except for the number 9,999,999,999 which it predicted completely correctly, all the predictions by ChatGPT3 were even worse than the classical ML models. This further showcases the deceptiveness of the simplicity of the task. The neural networks, on the other hand, produced the following results after the classification modification:
- 6 - Digit Dataset:

- Input: (999999, 999998) and (100000, 100001)
- Neural Network output: [0,0,0,0,0,0,0,0,0,5] and [0,0,0,0,0,0,0,0,1,4] for the former pair, and [5,1,0,0,0,0,0,0,0,0] and [4,2,0,0,0,0,0,0,0,0] for the latter.
- Neural Network with Embedding output: [0,0,0,0,0,0,0,0,0,6] and [0,0,0,0,0,0,0,0,1,5] for the former pair, and [5,3,0,0,0,0,0,0,0,0] and [3,3,0,0,0,0,0,0,0,0] for the latter.
- 10 - Digit Dataset:
  - Input: (9999999999, 9999999998) and (1000000000,1000000001)
  - Neural Network output: [0,0,0,0,1,1,0,0,0,9] and [0,0,0,0,1,1,0,0,1,8] for the former pair, and [7,2,1,0,1,0,0,1,0,0] and [7,2,1,0,0,0,1,0,0,0] for the latter.
  - Neural Network with Embedding output: [0,1,0,0,0,1,0,2,2,9] and [0,0,0,0,1,0,0,0,1,9] for the former pair, and [9,1,0,0,0,1,0,0,0,0] and [9,2,0,0,0,0,2,0,0,1] for the latter.

Interestingly, half of these predictions are incorrect but the other half are either completely correct or close to it with one or so digits wrong. They, at least, do not make the exact same prediction for the successive numbers unlike the classical ML models which means that they are partially learning the pattern. However, similar to classical ML models, their performance significantly worsens for 10-digit numbers as well. The proportion of data seems to play a significant role in the performance of all the models but with varying degrees.

*Ablation Study*

When running the neural networks on the 6-digit and 10-digit test sets, we found some alternative hyperparameter values, learning rate (lr) and layers, which gave significantly better outputs in terms of the regression metrics. We have mentioned them in the table given below, refer to Table 5.

**Table 5**. Alternative hyperparameter values for neural networks on the test sets

| 6-Digit Test Set | Hyperparameters | RMSE | MAE |
| --- | --- | --- | --- |
| Neural Network + Embedding | lr = 1e-5, layers = (96,96,96) | 0.093 | 0.073 |
| **10-Digit Test Set** | | | |
| Neural Network | lr = 0.003, layers = [128,128,128] | 0.171 | 0.130 |
| Neural Network + Embedding | lr=5e-3, layers = [256,256,256] | 0.221 | 0.168 |

## Conclusion

In this research work we compared the performance of different classical machine learning models and neural networks in identifying the frequency of occurrences of each digit in a given number. We observed that the neural networks significantly outperformed the classical ML models in terms of both the regression and classification metrics for both the 6-digit and 10-digit number datasets.

We discovered that some of the behaviors of the classical machine learning models such as split condition and averaging made the trees extremely biased and led to overfitting and memorization. Thus they failed in pattern recognition. The neural networks, on the other hand, thanks to their non-linear optimization were substantially more successful in recognizing the evident pattern. The accuracy was greater than 95% for all scenarios which indicates that the deep learning models did, in fact, learn the pattern accurately. This research further acknowledges the vast learning capabilities and adaptability of neural networks that have been stated in previous research work.

All the experiments were conducted on a MacBook M2 Air in a matter of two months. With more time, one could potentially extend the research to other datasets with larger numbers of digits and may find various other trends with neural networks. Regardless, they already seem to be reliable in learning this unconventional, yet simple pattern.
Furthermore, despite the research being experimental in nature, the results obtained in this research can potentially be applied to downstream computer vision problems, such as counting the number of times a specific object occurs in an image, which is an essential task in many computer vision applications[3,5,15,16]. Also, the ability to detect the most frequent elements can be used to detect the rare elements, which can have applications in healthcare, e.g. to detect rare diseases.

## Acknowledgement

I would like to acknowledge the unconditional support and guidance offered from my mentor Mr Viveka Kulharia, PhD in Computer Vision from the University of Oxford, for assisting me in everything, from researching the idea through his resources to writing the paper.